%% file: main.tex
\documentclass{article}

\usepackage[square,numbers]{natbib}
\usepackage[final]{neurips_2021}

\usepackage[utf8]{inputenc} 
\usepackage[T1]{fontenc}    
\usepackage{hyperref}       
\usepackage{url}            
\usepackage{booktabs}       
\usepackage{amsfonts}       
\usepackage{nicefrac}       
\usepackage{microtype}      
\usepackage{xcolor}         
\usepackage{graphics} 
\usepackage{epsfig} 
\usepackage{mathptmx} 
\usepackage{times} 
\usepackage{amsmath} 
\usepackage{amssymb}  
\usepackage{algorithm} 
\usepackage{float}
\usepackage{algpseudocode} 
\usepackage{multicol}
\usepackage{graphicx}
\usepackage{comment}
\usepackage{pgfplots}
\usepackage{tikz}
\usepackage{tikzscale}
\usepgfplotslibrary{groupplots}
\usepackage{svg}
\usepackage{siunitx}
\usepackage{cleveref}
\usepackage[font=small]{caption}
\usepackage[tight]{subfigure}
\usepackage{subfloat}
\usepackage{xr}
\hypersetup{
    urlcolor=black,
}

\crefname{figure}{Fig.}{Fig.}
\Crefname{figure}{Figure}{Figures}
\crefname{equation}{Eq.}{Eq.}
\Crefname{equation}{Equation}{Equations}

\newcommand{\ie}{\textit{i}.\textit{e}., }
\newcommand{\eg}{\textit{e}.\textit{g}., }
\newcommand{\st}{\text{s.t. }}

\DeclareMathOperator*{\argmin}{arg\,min}

\title{Data Efficient Human Intention Prediction: Leveraging Neural Network Verification and Expert Guidance}

\author{
    Ruixuan Liu\\
  Robotics Institute\\
  Carnegie Mellon University\\
  Pittsburgh, PA 15213 \\
  \texttt{ruixuanl@andrew.cmu.edu} \\
  \And
  Changliu Liu\thanks{This work is in part supported by Ford Motor Company. } \\
  Robotics Institute\\
  Carnegie Mellon University\\
  Pittsburgh, PA 15213 \\
  \texttt{cliu6@andrew.cmu.edu} 
}

\begin{document}

\maketitle

\begin{abstract}
Predicting human intention is critical to facilitating safe and efficient human-robot collaboration (HRC). However, it is challenging to build data-driven models for human intention prediction. One major challenge is due to the diversity and noise in human motion data. It is expensive to collect a massive motion dataset that comprehensively covers all possible scenarios, which leads to the scarcity of human motion data in certain scenarios, and therefore, causes difficulties in constructing robust and reliable intention predictors. To address the challenge, this paper proposes an iterative adversarial data augmentation (IADA) framework to learn neural network models from an insufficient amount of training data. The method uses neural network verification to identify the most ``confusing'' input samples and leverages expert guidance to safely and iteratively augment the training data with these samples. The proposed framework is applied to collected human datasets. The experiments demonstrate that our method can achieve more robust and accurate prediction performance compared to existing training methods. 
\end{abstract}

\section{INTRODUCTION}\label{intro}
\begin{figure}
\centering
\subfigure[]{\includegraphics[width=0.32\linewidth]{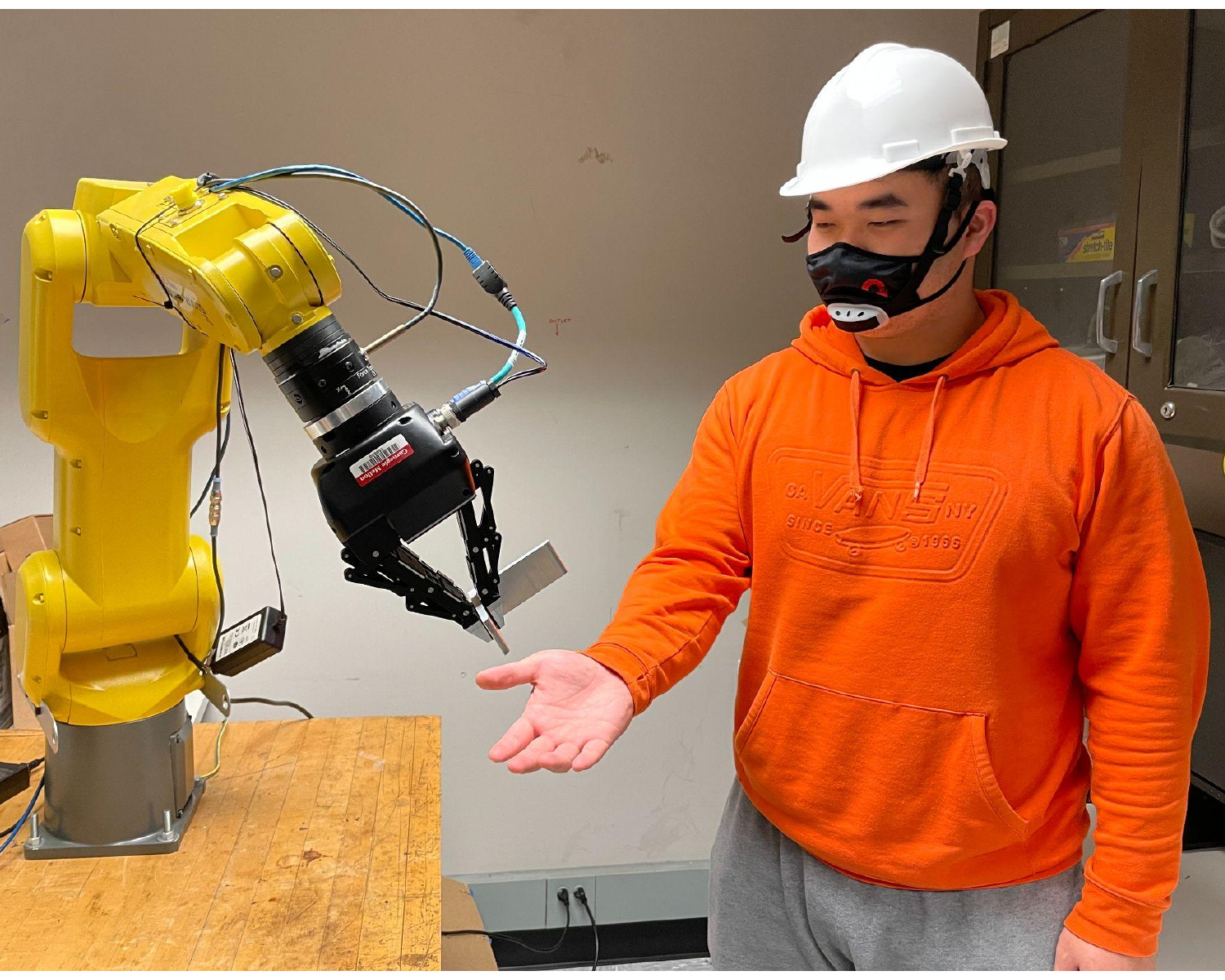}\label{fig:handover}}\hfill
\subfigure[]{\includegraphics[width=0.32\linewidth]{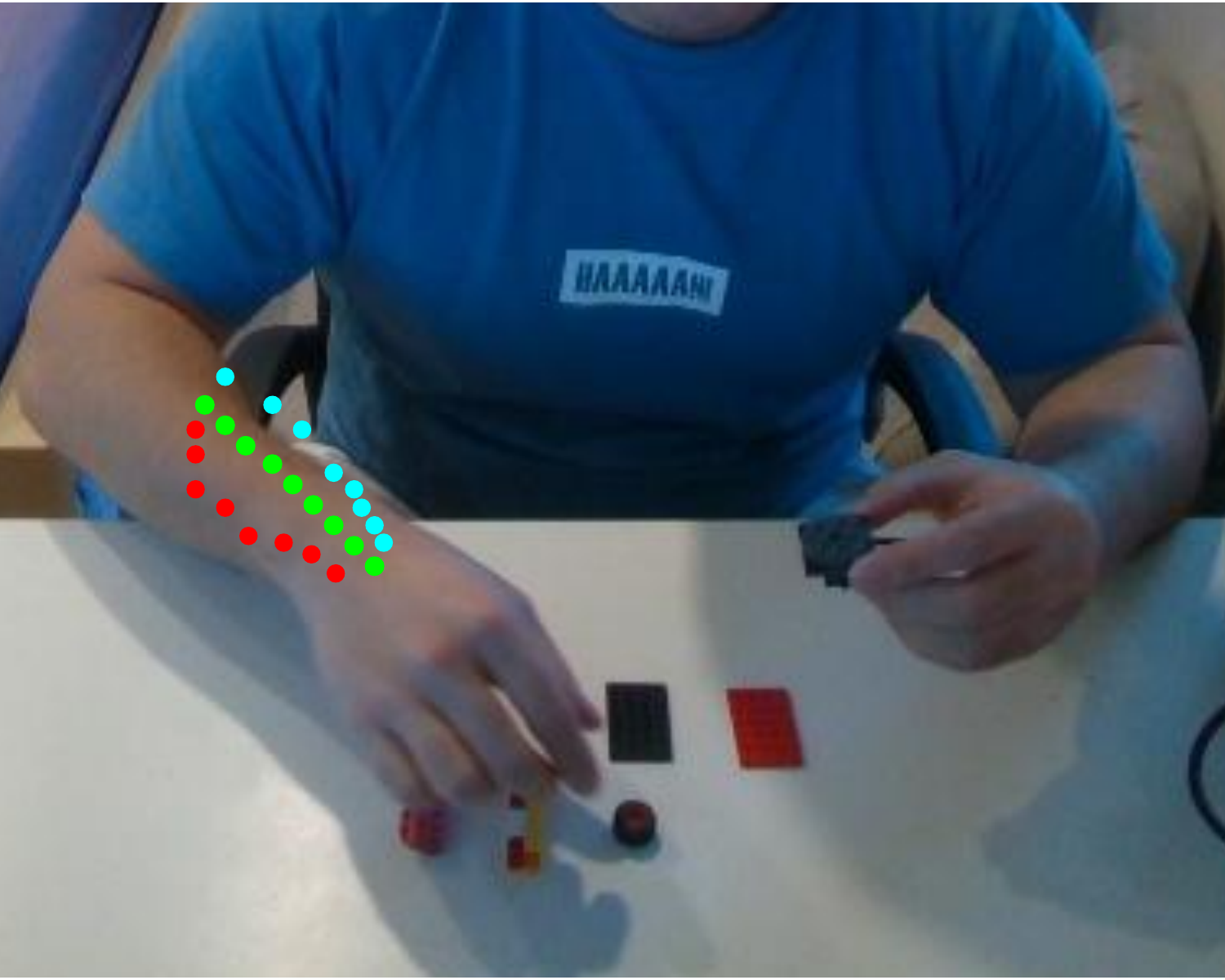}\label{fig:diff_motion}}\hfill
\subfigure[]{\includegraphics[width=0.32\linewidth]{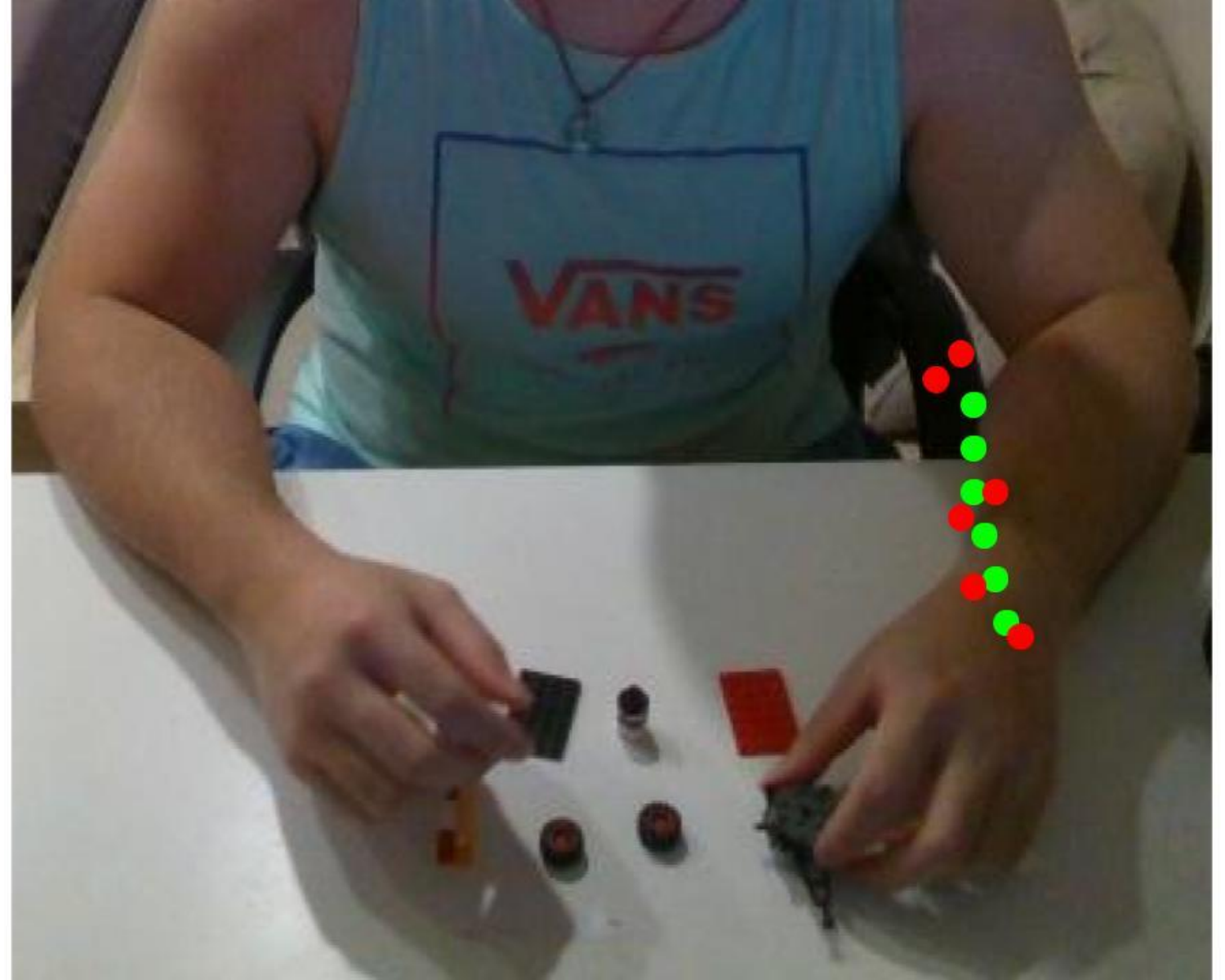}\label{fig:traj_noise}}
\vspace{-10pt}
\caption{(a) An example of HRC: Human-robot handover \cite{jpc,jssa}. (b) For the same type of motion, the human can perform in different ways as shown by the red, green and blue trajectories. (c) Noise in human motion data. Green: true wrist trajectory. Red: captured wrist trajectory.
\label{fig:adversarial_data_augmented}}
\vspace{-10pt}
\end{figure}

The rapid development of human-robot collaboration (HRC) addresses contemporary needs by enabling more efficient and flexible production lines \cite{Matheson_2019, VILLANI2018248}. Due to the close physical interactions (\cref{fig:handover}), any collision could lead to severe harm to the human workers. Therefore, it is essential to ensure safety while maximizing efficiency when facilitating HRC.
One key technology to enable safe and efficient HRC is to have accurate and robust human intention prediction \cite{book}.
Data-driven methods, especially neural network (NN) models, are widely used for intention prediction \cite{abu_intention, rudenko,8525781,7995948}.

This paper mainly focuses on human upper-body motion during HRC in manufacturing settings.
It is challenging to build data-driven models for intention prediction in such environments due to data scarcity. Since the prediction task varies when the HRC task changes (which happens frequently), it is expensive, if not impossible, to collect a dataset from different human subjects that comprehensively covers all possible scenarios. 
One reason that leads to the high cost is the diversity of human behavior. 
For example, in a setting where the humans are doing assembly tasks, human subjects with different habits, body structures, moods, and task proficiencies may exhibit different motion patterns. As shown in\cref{fig:diff_motion}, for the same reaching motion, the wrist can move in different ways. It is expensive (if not impossible) to collect data from all human subjects in all possible task situations. 
Another reason for the high cost of data collection is the existence of exogenous input disturbances. For example in \cref{fig:traj_noise}, due to the inevitable sensor noise and algorithm uncertainty, the captured motion trajectory (red) may deviate from the ground truth motion (green). It is expensive to generate a full distribution of these disturbances on each data point.

Due to these high costs of data collection, a well-distributed and sufficient human behavior dataset is usually not available. 
Early works \cite{286891, Japkowicz02theclass, 5128907, krawczyk2016learning} have shown that an insufficient amount of training data would make it difficult to train an NN model, as the performance of the learned model will deteriorate in testing. Therefore, the learned model is not deployable in real applications.
To deal with data deficiency, some methods \cite{9281312, 8814980} use online adaptation to incrementally update the NN using the data received online, which has been shown to improve prediction accuracy. However, these methods are post-deployment measures and there is no control over the incoming data, which might lead to safety hazards during the adaptation process. This paper investigates methods that efficiently learn NN models before deployment with limited data. The goal is to actively and cost-effectively augment the dataset (i.e., getting full control over the augmented data) so that the learned NN can perform robustly in real applications.

To achieve the goal, this paper proposes an iterative adversarial data augmentation (IADA) framework to train NNs with limited data. This framework leverages neural network verification and expert guidance for iterative data augmentation during training.
The key idea of IADA is to find the most ``confusing'' samples to the NN, \eg the samples on the decision boundary of the current NN, and add them back to the dataset to improve model accuracy. We use NN verification to find these samples by computing the closest adversaries to existing data (called roots) in $L_{\infty}$ norm. These samples are called ``adversaries'' since the current NN predicts that they have different labels from the labels of roots (hence they are on the decision boundary of the current NN). The IADA framework will query experts to label these samples in order to ensure the correctness of the labels of the adversaries. The sample is a true adversary if its ground truth label is the same as the label of its root; otherwise, this sample is a false adversary. This paper incorporates human-in-the-loop verification as the expert guidance to label the adversaries.
The labeled samples will be added back to the training dataset no matter what labels they are assigned. The true adversarial samples can improve the robustness of the network, while the false adversarial samples can help recover the ground truth decision boundary.
The IADA framework iteratively expands the dataset and trains the NN model.
To verify the effectiveness of IADA, we applied it to two human intention prediction tasks. We compared our training framework against several other training methods. The results demonstrate that our training method can improve the model testing accuracy by over $10\%$.

\section{Related Work}\label{relate}
\paragraph{Active Learning}
Active learning \cite{settles2009active} studies techniques to learn NN models from a \emph{limited} and \emph{partially labeled} training data with the guidance from human experts. 
Early works \cite{settles2009active,10.1023/A:1022673506211} select samples from a pool of unlabeled dataset and decide whether to query for their labels according to a designed query strategy, \ie uncertainty sampling \cite{10.1007/978-1-4471-2099-5_1}, query-by-committee \cite{10.1145/130385.130417}, expected model change \cite{NIPS2007_a1519de5}, etc. 
Recent work \cite{Wang_2017} improves the traditional uncertainty sampling with pseudo-labeling. 
\cite{ducoffe2018adversarial} proposes to use DeepFool \cite{moosavidezfooli2016deepfool}, an adversarial attack method, to find adversaries and then queries the labels of the unlabeled samples in the given training set that are close to the adversaries. In addition, it pseudo-labels the adversaries (assigns the same labels to the adversaries as the queried samples) in order to double the amount of augmented data.
\cite{Mayer_2020_WACV} uses a generative adversarial network (GAN) to generate adversaries. The learner queries the expert for the unlabeled training data that is the most similar to the adversary. 
However, most existing active learning frameworks assume unlabeled data is free, and they only query the labels of the unlabeled samples from the given dataset, which is constrained by a discrete input space.
On the other hand, this paper tackles the scenario that additional data is as expensive to collect as the training data (\ie training data is fully labeled but the amount is limited). Therefore, it is important that the framework can generate informative data samples in the entire input space.  

The proposed method in this paper is more closely related to membership query synthesis (MQS) \cite{angluin1988queries,ANGLUIN2004175}, which is a category in active learning. MQS queries the labels as well as generates the data to be queried. 
Early works generate new data by random sampling. 
\cite{zhu2017generative} proposes to use a pre-trained GAN to generate new data and query the oracle for labels. However, the GAN has poor explainability, which might lead to safety concerns.
How to generate informative data remains questionable.

\paragraph{Data Augmentation}
Learning from insufficient training data has been widely studied recently \cite{286891, Japkowicz02theclass, 5128907, krawczyk2016learning}. Data augmentation (DA) is a widely-used approach. People use different approaches to generate additional data given the existing dataset to improve the generalizability of the classifiers \cite{Shorten2019ASO}.
The augmented data is assumed to have an identical label as the root data (\ie data that the augmented data is generated from).
\cite{cutout,mixup,zhong2017random,da_gaussian,augmix, cubuk2019autoaugment,Zhang2020AdversarialA} propose either to manually design a policy or search for an optimal policy among the pre-defined policies to generate new data.
However, manually designing the augmentation policy requires strong domain knowledge.
In addition, the designed policy might only be suitable for a small range of related tasks and whether the policy is optimal remains questionable.
On the other hand, instead of explicitly designing the policy, \cite{perez2017effectiveness, Lemley_2017, antoniou2018data, bowles2018gan} propose to learn generative NNs (\eg GAN) to create new data.
Using GANs for DA is knowledge-free. However, it has poor explainability, which might be a potential concern for safety-critical tasks similar to the ones used in MQS.
Recent works \cite{deepaug,Zhang2020AdversarialA,advaug,maxup} propose to augment the data using adversarials.
Nevertheless, how to find informative adversarial samples still remains a question.
In addition, existing works follow the definition in \cite{NEURIPS2018_8562ae5e} that adversarial examples have the same labels as the root data, which is applicable to image-related tasks. Since image data usually has trivial distribution on the class boundaries, the adversarial perturbation can hardly change the true label of an image \cite{goodfellow2015explaining} in most tasks.
However, it is difficult to apply to our setting (\ie intention prediction) since it is easy to incorrectly take false adversaries as true adversaries (\ie trajectories can be easily perturbed to change the true labels), which might harm the model training.

\paragraph{Adversarial Training}
Adversarial training \cite{bai2021recent} intends to learn NN models robust to unseen adversarial data. 
It has been widely studied since \cite{szegedy2014intriguing} first introduced the adversarial instability of NNs. 
Given the existing training data $D_0$, the adversarial training objective is formulated as a minimax problem
\begin{equation}\label{eq: adversarial training}
    \begin{split}
        \min_{\theta}\mathbf{E}_{(x_i, y_i)\in D_0}{\max_{||\delta_i||_{\infty}\leq \epsilon} L(f_{\theta}(x_i+\delta_i), y_i)},
    \end{split}
\end{equation}
where $\delta_i$ is the adversarial perturbation and $\epsilon$ is the maximum allowable perturbation. Early works \cite{szegedy2014intriguing, goodfellow2015explaining} efficiently estimate the adversarial perturbations based on the gradient direction. However, \cite{moosavidezfooli2016deepfool} showed that the estimation in the gradient direction might be inaccurate, and thus, makes the trained model sub-optimal. Recent work \cite{cheng2020cat} proposes to adaptively adjust the adversarial step size during the learning. Based on the idea of curriculum learning, \cite{zhang2020attacks} proposes to replace the traditional inner maximization with a minimization, which finds the adversarial data that minimizes the loss but has different labels. The adversarial data generated via the minimization is also known as the \textit{friendly adversarial data}. 

Although it has been shown that adversarial training can improve NN robustness against adversarial attacks, whether it can maintain the performance on clean data is unclear since it might mix true and false adversaries as mentioned in adversarial data augmentation.
Moreover, adversarial training only improves the local robustness around the training data, and thus, has limited capacity to improve the network performance if the training data is scarce.  

\section{Problem Formulation} \label{PF}
This paper focuses on human intention prediction, which is essentially a classification problem.
At timestep $k$, the prediction model is given with the human and environment information $x_k\in X$, (\eg historical human motion, human pose, obstacles poses, etc), where $X$ is the input space, and outputs the behavior label $y$. 
Note that depending on the human task, the input and output could encode different information.
The goal is to construct a prediction model $f(\cdot)$ that maps the human and environment observation to the corresponding behavior label, \ie $y=f(x)$. This paper parameterize $f(\cdot)$ using a neural network with parameter $\theta$.
However, due to data scarcity, it is difficult to learn a robust and accurate $f(\cdot)$.

Given a training dataset $D_0=\{(x_i, y_i)\}_{i=1}^n$ where $n$ is the size of $D_0$, 
the regular supervised training learns the model by solving the following optimization
\begin{equation}\label{eq:reg_training_d0}
    \argmin_{\theta}\frac{1}{n}\Sigma_{i=1}^n{L(f_{\theta}(x_i), y_i)} \left(=:\mathbf{E}_{(x_i, y_i)\in D_0}L(f_{\theta}(x_i), y_i)\right),
\end{equation} 
where $L(\cdot)$ is the loss function, 
and $f_{\theta}(\cdot)$ is the NN transfer function. 
However, the true goal in real applications is to learn a model that minimizes the expected loss when deployed in real applications:  
\begin{equation}\label{eq: ground truth theta}
    \argmin_{\theta}\mathbf{E}_{(x, y)\in D}(L(f_{\theta}(x),y)),
\end{equation}
where $D$ represents the real input-output data distribution, which is unavailable during training. When $D_0$ is similar to $D$, we can obtain a $f_{\theta}(\cdot)$ that behaves similarly as the ground truth $f(\cdot)$. However, in our case (\ie intention prediction), $D_0$ is limited and biased, thus, leading to a poorly trained $f_{\theta}(\cdot)$ that behaves differently as $f(\cdot)$. The goal is to learn a robust and accurate $f_{\theta}(\cdot)$ offline given the insufficient human dataset.

\section{IADA: Iterative Adversarial Data Augmentation} \label{approach}

Due to the data scarcity, an NN classifier can achieve outstanding performance on the training data but fails to perform well on the testing data or real deployment. 
To address the challenge, we aim to augment the insufficient dataset $D_0$ in \cref{eq:reg_training_d0} with adversarial data $D_0'\cup D_{adv}$, where $D_0'$ is more trustworthy than $D_{adv}$, to robustly learn the true decision boundaries of the real but unknown data $D$ as discussed in \cref{eq: ground truth theta}. 
The goal is to ensure the learned model on $D_0\cup D_0'\cup D_{adv}$ is as close as possible to \cref{eq: ground truth theta}. 
To achieve the goal, we formulate the training problem as a two-layer optimization
\begin{equation}\label{eq:IADA_prob}
\begin{split}
    \argmin_{\theta}~&\mathbf{E}_{(x, y)\in D_0\cup D_0'\cup D_{adv}}(L(f_{\theta}(x),y)),\\
    & \begin{Bmatrix}
    D_{adv}\\
    D_0'
    \end{Bmatrix}=\Bigg\{\begin{matrix}
    \left(x', \text{label}_a(x')\right)\\
    \left(x', \text{label}_e(x')\right)
    \end{matrix}\mid \exists (x_0,y_0)\in D_0\cup D_0', x'=\arg\min_{x'}L'(f_{\theta}(x'), y_0)\\
    &~~~~~~~~~~~~~~~~~~~~~~~~~~~~~~~~~~~~~~~~~~~~~~~~~~~~~~~~~~~~~~~~~~~~~~~~~\st ||x_0-x'||_{\infty}\leq\epsilon \wedge f_{\theta}(x') \neq y_0 \wedge x'\in X \Bigg\}.
\end{split}
\end{equation}
The outer objective optimizes the NN model parameters $\theta$ to minimize the loss of the data from $D_0\cup D_0' \cup D_{adv}$.
The inner objective constructs $D_0' \cup D_{adv}$ given $\theta$, which essentially finds the most friendly adversarial data for the training data. The most friendly adversarial data for $(x_0,y_0)\in D_0\cup D_0'$ is an input sample $x'$ that is at most $\epsilon$ distance away from $x_0$ in the $L_\infty$ norm but changes the network output from $y_0$ with the smallest loss $L'(\cdot)$. The loss $L'(\cdot)$ for the inner objective may or may not be the same as the original loss $L(\cdot)$. Ideally, we should choose a loss that guides us to the most ``confusing'' part of the input space.
The label of these friendly adversarial data is decided by additional labeling functions, $\text{label}_a(\cdot)$ (auto labeling) and $\text{label}_e(\cdot)$ (expert labeling), which ideally should match the ground truth. If $\text{label}(x')=y_0$, we call $x'$ a true adversary; otherwise a false adversary.

\begin{figure}
    \centering
    \includegraphics[width=\linewidth]{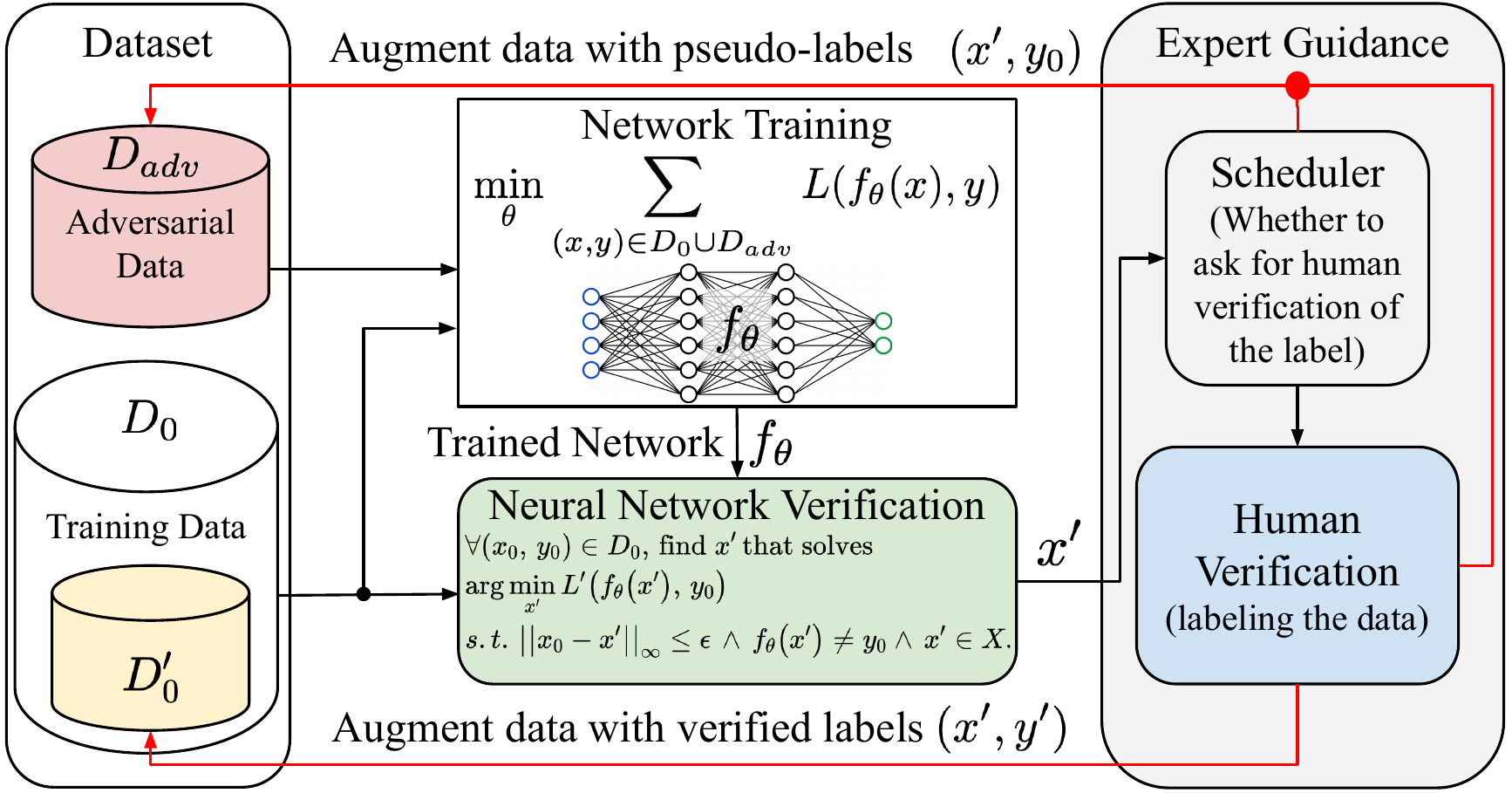}
    \caption{Iterative Adversarial Data Augmentation (IADA) training framework. For $(x,y)\in D_0\cup D_0' \cup D_{adv}$, the framework continuously updates the NN model. If $(x_0,y_0) \in D_0 \cup D_0'$, the neural network verification finds the most friendly adversarial sample $x'$ around $x_0$ if exists. The scheduler determines if human verification is required for $x'$. If not, $(x', y_0)$ is augmented to $D_{adv}$. Otherwise, if human is able to assign $y'$, then $(x',y')$ is added to $D_0'$. If not, then $(x',y_0)$ is appended to $D_{adv}$. \label{fig:IADA}}
    \vspace{-10pt}
\end{figure}

We propose an iterative adversarial data augmentation (IADA) training framework to solve \cref{eq:IADA_prob}. 
\Cref{fig:IADA} illustrates the proposed training framework that aims to solve the problem defined in \cref{eq:IADA_prob}. In particular, the inner objective will be solved using neural network verification to be discussed in \cref{VAT}, while the labeling will be performed under expert guidance to be discussed in \cref{HV}. The iterative approach to solve the two-layer optimization will be discussed in \cref{VDA}.

\subsection{Neural Network Verification to Find Adversaries} \label{VAT}
The two-layer optimization in \eqref{eq:IADA_prob} is using adversarial data for data augmentation. Unlike the conventional minimax formulation of adversarial training \eqref{eq: adversarial training}, the major distinction lies in the inner objective. 
The minimax formulation in \eqref{eq: adversarial training} might reduce the model accuracy and generalizability when maximizing the robustness \cite{tsipras2019robustness}.
Similar to the approach in \cite{zhang2020attacks}, instead of finding the adversaries that maximize the loss in \cref{eq: adversarial training}, we 
find the ``closest'' adversaries by designing the inner loss to penalize the magnitude of $\delta_0$. Then the inner optimization in \eqref{eq:IADA_prob} can be written as
\begin{equation}\label{eq:verification_obj}
    \argmin_{x'}{\|x_0 -x'\|_{\infty}}, ~ \st ~ {\|x_0 -x'\|_{\infty}\leq \epsilon,~ f_{\theta}(x_0) \neq f_{\theta}(x')},~ x'\in X.
\end{equation}
The reason why we use the $L_{\infty}$ distance metric in the input space as the loss $L'(\cdot)$ instead of using the original loss or any other loss that penalizes the output is that this metric reflects the ``confusing'' level of samples. We generally expect that the learned NN model is regular and robust in the training data for generalizability. The easier it is to change the label by perturbing the input data, the less regular the model is, and hence more ``confusing''. In addition, this formulation provides a quantitative metric $\delta_0$ for evaluating the robustness online during training. Therefore, we can prioritize enhancing the weaker boundaries and obtain full control of the training process. 

Nevertheless, the optimization in \eqref{eq:verification_obj} is nontrivial to solve due to the nonlinear and nonconvex constraints introduced by $f_\theta(\cdot)$. To obtain a feasible and optimal solution, we use neural network verification algorithms \cite{liu2020algorithms} to find the appropriate $x'$.
Note that this paper assumes that the NN models only contain ReLU activations \cite{relu}. 
The optimization in \eqref{eq:verification_obj} essentially finds the minimum input adversarial bound, which can be solved by various neural verification algorithms. In particular, primal optimization-based methods such as MIPVerify \cite{tjeng2019evaluating} and NSVerify \cite{lomuscio2017approach} solve the problem exactly by encoding the neural network into a mixed-integer linear program; dual optimization-based methods \cite{convdual} can compute an upper bound of the problem by relaxing the nonlinear ReLU activation functions; reachability-based methods \cite{weng2018fast} can also compute an upper bound by over-approximating the reachable set and binary search for the optimal loss.
Based on the computation need and the accuracy requirement, \cref{eq:verification_obj} can be solved using different methods.

\subsection{Expert Guidance for Labeling}\label{HV}
With the adversaries $x'$ obtained from neural network verification, we need to label them before augmentation. 
The proposed IADA framework uses expert guidance to guarantee the safety and accuracy of the training. 
Note that the underlying assumption is that the human oracle has ground-truth knowledge.
The framework queries human experts to verify the newly added adversaries. 
It uses a scheduler to balance the required human effort and the training performance. 
We use the $L_{\infty}$ distance check to construct the scheduler since it indicates the distance between the adversarial data $x'$ and the root $x_0$.
In general, a smaller distance value indicates the two data points are more similar and more likely to share the same label, whereas a larger distance represents the data points are different.
The scheduler requires human verification if $||x_0-x'||_{\infty}>d$, where $d$ is a pre-defined threshold.
The unverified adversaries will be pseudo-labeled by assigning the identical label as the root.
Based on the application, a smaller $d$ can be chosen to require more human effort to improve the training and data expansion performance, while a larger $d$ alleviates the amount of human effort. 

\subsection{Iterative Adversarial Data Augmentation}\label{VDA}
We use an iterative approach to solve the two-layer optimization in \eqref{eq:IADA_prob} by incrementally augmenting the data with the outer NN training loop. 
In one iteration, we find the adversaries of the data in $D_0\cup D_0'$ in \cref{VAT}, which are the most ``confusing'' points for the current NN. The adversaries will be labeled by an expert and augmented to the dataset, either $D_0'$ or $D_{adv}$, as shown in \cref{fig:IADA}. In the next iteration, the framework will further verify, label, and expand the dataset $D_0\cup D_0'\cup D_{adv}$. 

\begin{algorithm}[t]
    \caption{Iterative Adversarial Data Augmentation (IADA)\label{alg:IADA}} \small
	\begin{algorithmic}[1]
	    \State Input: Original dataset $(x_0, y_0) \in D_0$.
	    \State Input: Robustness bound $\epsilon$, inner optimization rate $r$, number of seeds $C$, learning rate $\alpha$.
	    \State Output: NN parameter $\theta$.
	    \State Initialize: $\theta = \theta_0$, $Q_v=\{D_0\}$.
	    \While{$Q_v$ not empty \&\& $k\leq max\_epoch$}
		    \If{InnerOptimizerRound($r$)}
		        \State Clear $D_{adv}$. \Comment{Refresh pseudo-labeled $D_{adv}$ since it might contain data with incorrect label.}
		        \For{$i=1,\dots, C$}
		            \State Breaks if $Q_v$ is empty.
		            \State $(x,~y) = Q_v.pop$. \Comment{Prioritize the point with 1) the lowest level of expansion and 2) the smallest perturbation bound.}
		            \State $x' = FindAdversary(\theta, x, y, \epsilon)$. \Comment{Find an adversary by solving \eqref{eq:verification_obj}.}
		            \State Skip if no adversary $x'$ is found.
		            \If{Scheduler \&\& Human Verified}
		                \State Obtain verified label $y'$. \Comment{Query human expert.}
		                \State Push $(x, y)$ and $(x', y')$ to $Q_v$ and append $(x', y')$ to $D_{0}'$.
		            \Else
		                \State Push $(x, y)$ to $Q_v$ and append $(x', y)$ to $D_{adv}$. \Comment{Pseudo-labeling.}
		            \EndIf
		        \EndFor
		    \EndIf
		    \For{minibatch $\{X,Y\}$ in $D_0 \cup D_0' \cup D_{adv}$}
	           \State $\theta \longleftarrow \theta-\alpha L(f_{\theta}(X), Y)$.
		    \EndFor
		\EndWhile
	\State \Return{$\theta$}.
	\end{algorithmic} 
\end{algorithm}

The framework maintains three datasets $D_0$, $D_0'$ and $D_{adv}$ as shown in \cref{fig:IADA}, where $D_0'$ and $D_{adv}$ are initially empty and $D_0$ is the initial training data. The adversaries are only augmented to $D_0'$ if they are verified by a human expert, otherwise, they are pushed into $D_{adv}$. Note that the adversaries generated in \cref{VAT} can either be true or false adversaries, but both are informative for improving the NN learning. However, incorrectly mixing these two types of adversaries can harm NN learning. Therefore, we maintain $D_0'$ and $D_{adv}$ to distinguish the safe data and potentially incorrect data.
The framework iteratively expands only from the data in $D_0\cup D_0'$, which are safe. Therefore, the framework can further expand the available training data safely.
$D_{adv}$ will be refreshed before neural verification as shown in \cref{alg:IADA}, since the framework only wants a temporary effect from $D_{adv}$ but a permanent effect from $D_0\cup D_0'$ for safety and correctness.
Note that the IADA framework can be viewed as an active learning method.
In addition, it is similar to standard data augmentation and adversarial training (but with iterative expansion) when the scheduler never requires human verification. It is, on the other hand, similar to data aggregation in imitation learning \cite{dagger} when the scheduler always queries for human knowledge.
Based on the applications, the scheduler can be tuned to either require more or less expert knowledge to balance the training performance and the human effort.

The IADA framework is summarized in \cref{alg:IADA}.
Due to the expensive computation needed for solving \cref{eq:verification_obj}, we introduce the inner optimization rate $r$ and the number of seeds $C$. $r$ indicates the frequency of online neural verification and data augmentation and $C$ indicates a maximum number of seeds to find adversaries at each inner optimization round. $Q_v$ is a priority queue based on the level of expansion and the perturbation bound $\delta$. The level is defined as 0 for the original training data. The adversaries generated from original data have level 1. $Q_v$ determines the weakest points and prioritizes them during verification and data expansion. In line 10, the system gets the weakest point and finds its adversarial sample on line 11. If an adversary is found, the system either queries for expert knowledge to assign a label or pseudo labels the adversary as discussed in \cref{HV}. If the label is assigned by the expert, the adversarial sample is added to $D_0'$ for further expansion, otherwise temporarily to $D_{adv}$. The NN model is updated using the training data and the adversarial data on lines 21-23.
Note that $r$ and $C$ in \cref{alg:IADA} will influence the performance of IADA. A more detailed study of these parameters is provided in the supplementary.

\subsection{IADA Analysis} \label{IADA_analysis}
The IADA framework is unique in several ways. First, it is well-known that there exists a trade-off between accuracy and robustness in general adversarial training \cite{tsipras2019robustness}. This is mainly due to incorrectly mixing the true and false adversaries. However, by introducing expert guidance, the true adversaries can improve the NN robustness and enlarge the decision area, while the false adversaries can enhance the ground truth decision boundary for better accuracy.
Second, although the augmented data may not recover the ground truth data distribution, the augmented data will converge to the ground truth decision boundary. As a result, the learned NN will converge to the ground truth with the correct decision boundary.
Based on these features, we argue that IADA is particularly suitable for tasks that have a non-trivial distribution of the data around the decision boundary, \ie small perturbation could have a significant impact on the model output, and with scarce training data.
And therefore, it is suitable to train intention classifiers using IADA.
Because data is usually expensive and small perturbation in the trajectories can easily change the intention, making it critical to distinguish the true and false adversaries.


\begin{figure}
\centering
\subfigure[]{\includegraphics[width=0.3\linewidth]{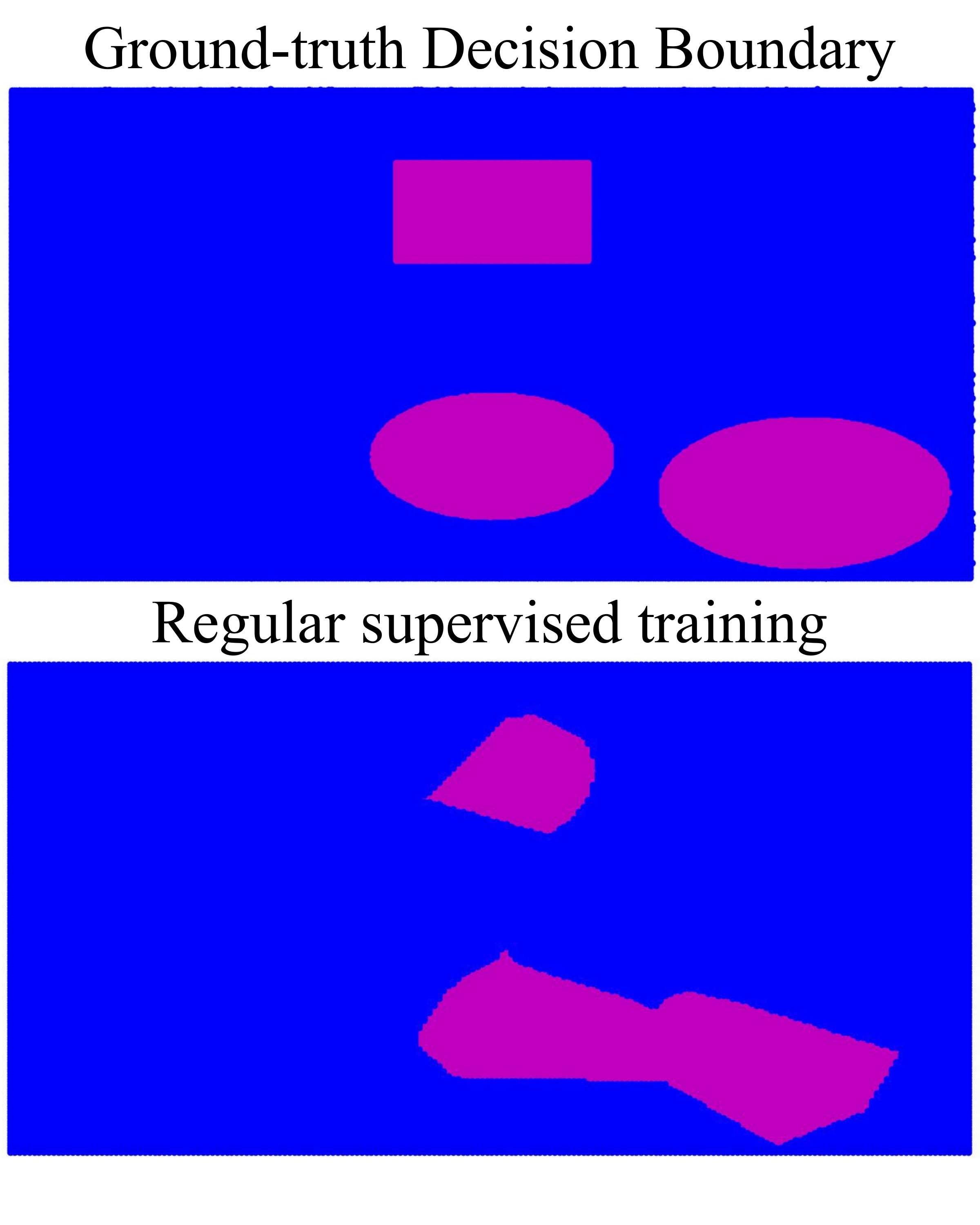}\label{fig:2d_data}}\hfill
\subfigure[]{\includegraphics[width=0.65\linewidth]{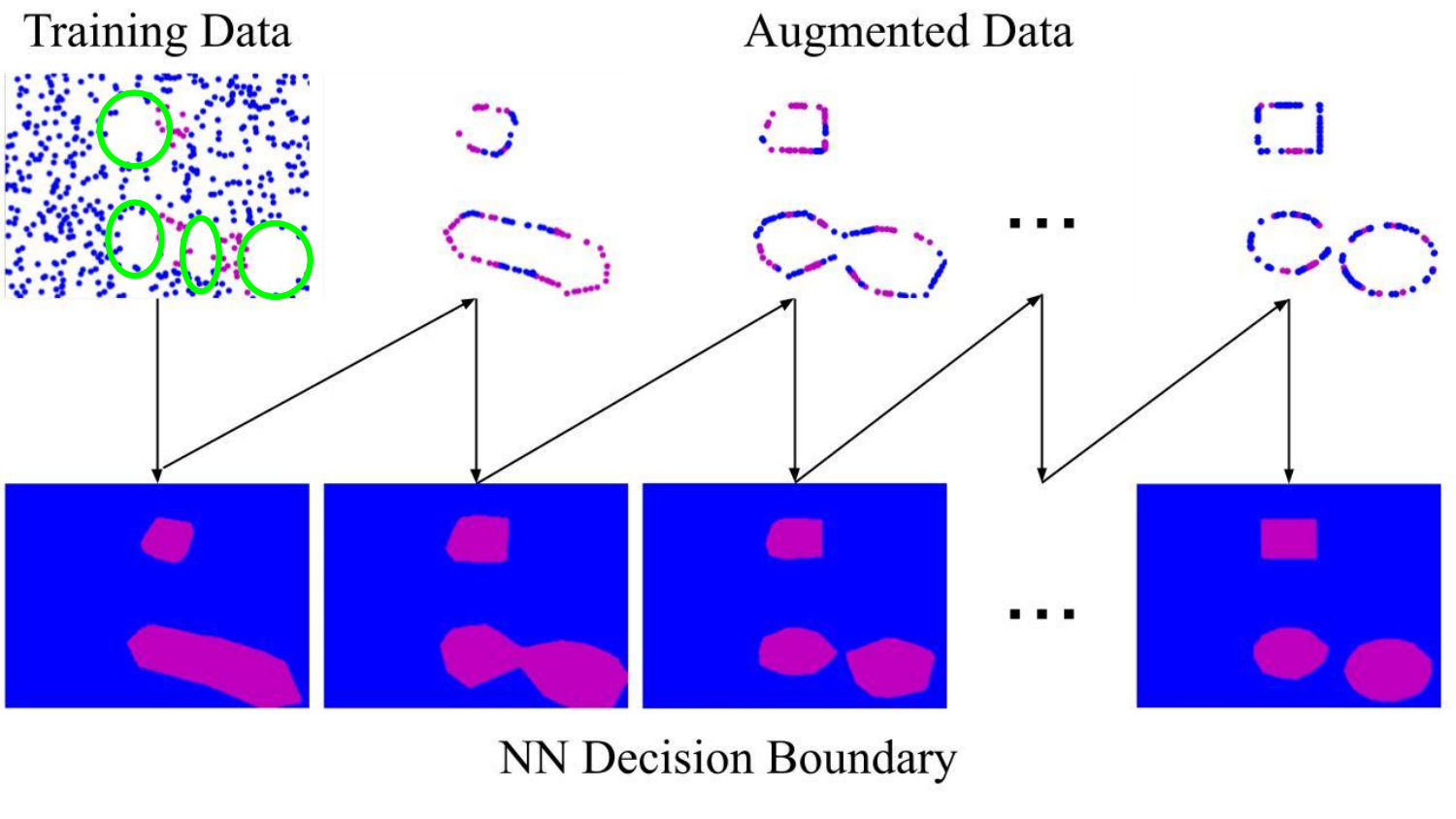}\label{fig:2d_training}}
\vspace{-10pt}
\caption{IADA training on a binary classification problem. (a) Binary classification data. Top: ground-truth input-output space. Bottom: NN classifier learned by regular supervised training. (b) IADA training process. Top-left: available training data. The green circles highlight the scarcity of the training data. First row: adversarial data augmented by neural network verification and expert guidance given the current NN model. Second row: learned NN decision boundary. Bottom-right: NN classifier learned by IADA.
\label{fig:2d_example}}
\vspace{-10pt}
\end{figure}

\Cref{fig:2d_example} illustrates an example of IADA training on a 2D binary classification task for better visualization and understanding of the training process. 
In this task, slight perturbation (either in the $x$ or $y$ axis) could easily change the label of a data point.
\Cref{fig:2d_data} shows the ground-truth space of the problem, where we have two classes, blue and purple. Given an available training dataset (\cref{fig:2d_training}), we want the NN classifier to recover the true decision boundary. 
Note that the sampled training data is scarce as shown in \cref{fig:2d_training}, in which the blue data points are sampled from the entire input space whereas the purple data points are sampled only from half of the rectangle and circles. 
As a result, the classifier learned using regular supervised training cannot correctly recover the three purple regions (\cref{fig:2d_data}).
On the other hand, \cref{fig:2d_training} shows the data augmentation process of IADA. The augmented data converges to the true decision boundary, leading the learned classifier to recover the true decision boundary.

\section{Experiment Results} \label{results}

We evaluate the proposed IADA framework by training NN classifiers in two different human intention prediction problems.
We compared three training methods for each problem, including the regular supervised training (REG), the robust training via the convex outer adversarial polytope (COAP) \cite{convdual}, and our IADA training. The regular and IADA training use the Cross-Entropy loss with the Adam optimizer \cite{kingma2017adam} implemented in PyTorch \cite{NEURIPS2019_9015}.
All experiments were run in Windows 10 with AMD Ryzen 3700X 8-Core processor, 16GB RAM, and an RTX 2070 Super GPU.

\subsection{Intention Prediction 1: Human Assembly}
We apply IADA to learn the intention prediction model in a human assembly task (\cref{fig:diff_motion} and \cref{fig:traj_noise}), where the human uses the LEGO pieces on the table to assemble target objects.
We follow the data collection in \cite{9281312} and use an Intel RealSense D415 to capture the human motion at 30 Hz. 
The OpenPose \cite{8765346} is used to extract human wrist poses.
The input to the classification model is the previous 10-step ($\sim$0.3s) historical trajectories for both right and left wrists and the output is an intention label, indicating either the human is assembling, retrieving a piece, reaching for a piece, or doing irrelevant tasks.
The classifier is trained using the first two trials of the human and tested using the third trial. 
The classifier is constructed using a single-layer FCNN with 32 hidden neurons due to the small scale of the problem.
The learning rate is set to $0.01$ for all methods and the $\epsilon$ in \cref{eq:verification_obj} is set to $0.05$ for COAP and IADA.
For IADA, the innter optimization rate is set to be $r = 500$, and we have $C=5000$ and $d=0.01$. 
The neural network verification is implemented using the NeuralVerification.jl toolbox \cite{Liu2019iclr}. 

\begin{table}
\centering
\vspace{-10pt}
\caption{Comparison of the intention prediction accuracy in the human assembly task.} \label{Intentiontable1}
\begin{tabular}{c | c  c  c} 
\hline
 & REG & COAP\cite{convdual} & IADA \\  
\hline
Epochs: 500 &  81.99\% & 77.94\% & \textbf{82.53\%} \\
Epochs: 1k &  \textbf{85.92\%} & 77.95\% & 85.48\% \\
Epochs: 2k &  85.26\% & 81.66\% & \textbf{88.75\%} \\
Epochs: 3k &  82.97\% & 83.95\% & \textbf{89.52\%} \\ 
Epochs: 4k & 82.86\% & 83.95\% & \textbf{90.83\%} \\ 
Epochs: 5k & 81.55\% & 83.95\% & \textbf{92.03\%} \\ 
\hline
\end{tabular}
\vspace{-10pt}
\end{table}

The model validation accuracy is shown in \cref{Intentiontable1}. 
We can see that the accuracy for REG and COAP increases initially. 
But soon the performance starts to drop in REG due to the overfitting while COAP reaches a steady state. 
On the other hand, by using expert knowledge and iterative expansion, IADA is able to continuously expand the data input space and improve the model accuracy. 
As we further increase the training epochs, we expect the prediction accuracy of the model learned by IADA will further increase.
More detailed visualizations are included in the supplementary.

\subsection{Intention Prediction 2: Human-robot Handover}
\begin{figure}
\centering
\subfigure[Working.]{\includegraphics[width=0.32\linewidth]{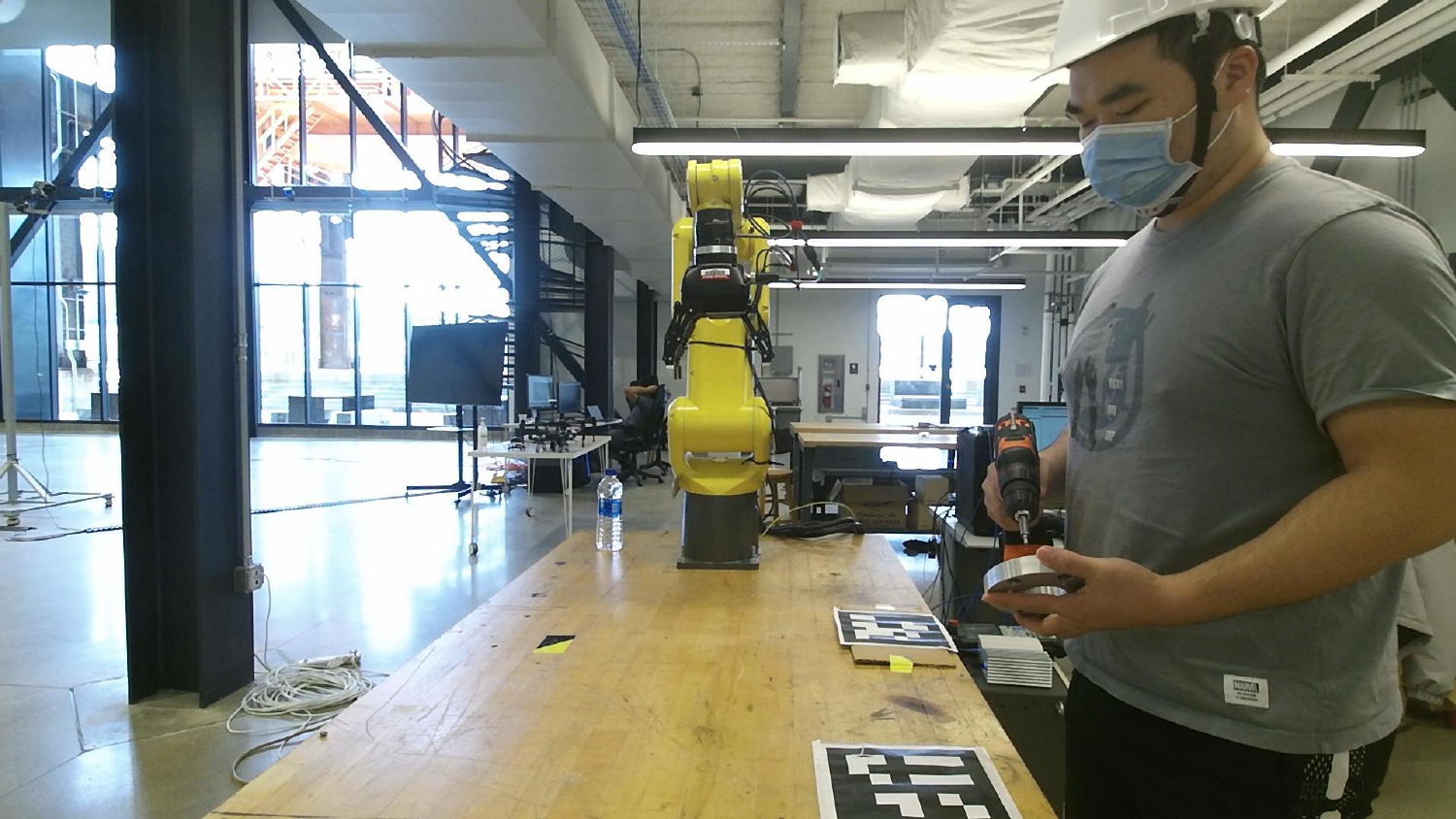}\label{fig:handover_img1}}\hfill
\subfigure[Requesting the power drill.]{\includegraphics[width=0.32\linewidth]{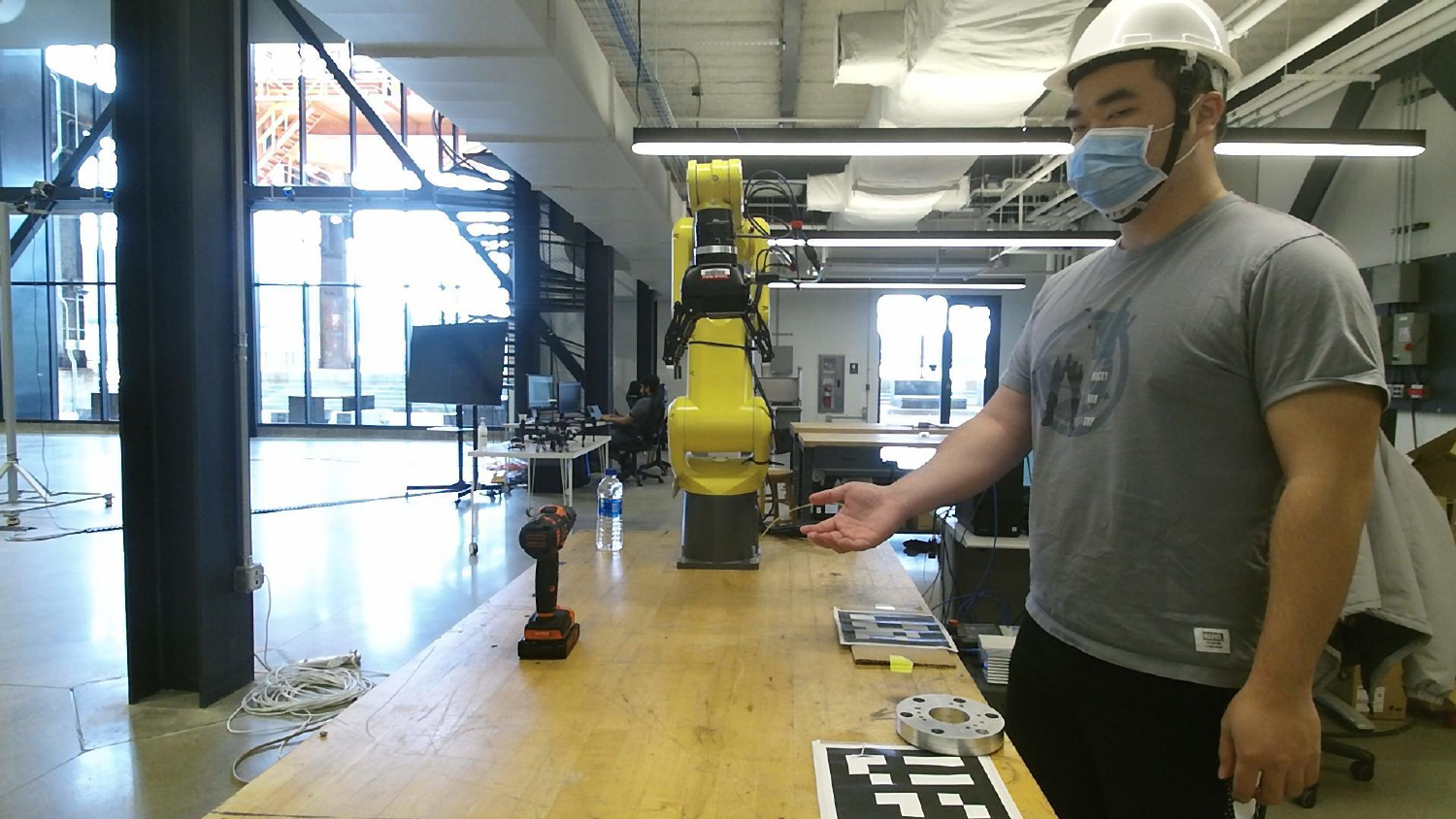}\label{fig:handover_img2}}\hfill
\subfigure[Relaxing.]{\includegraphics[width=0.32\linewidth]{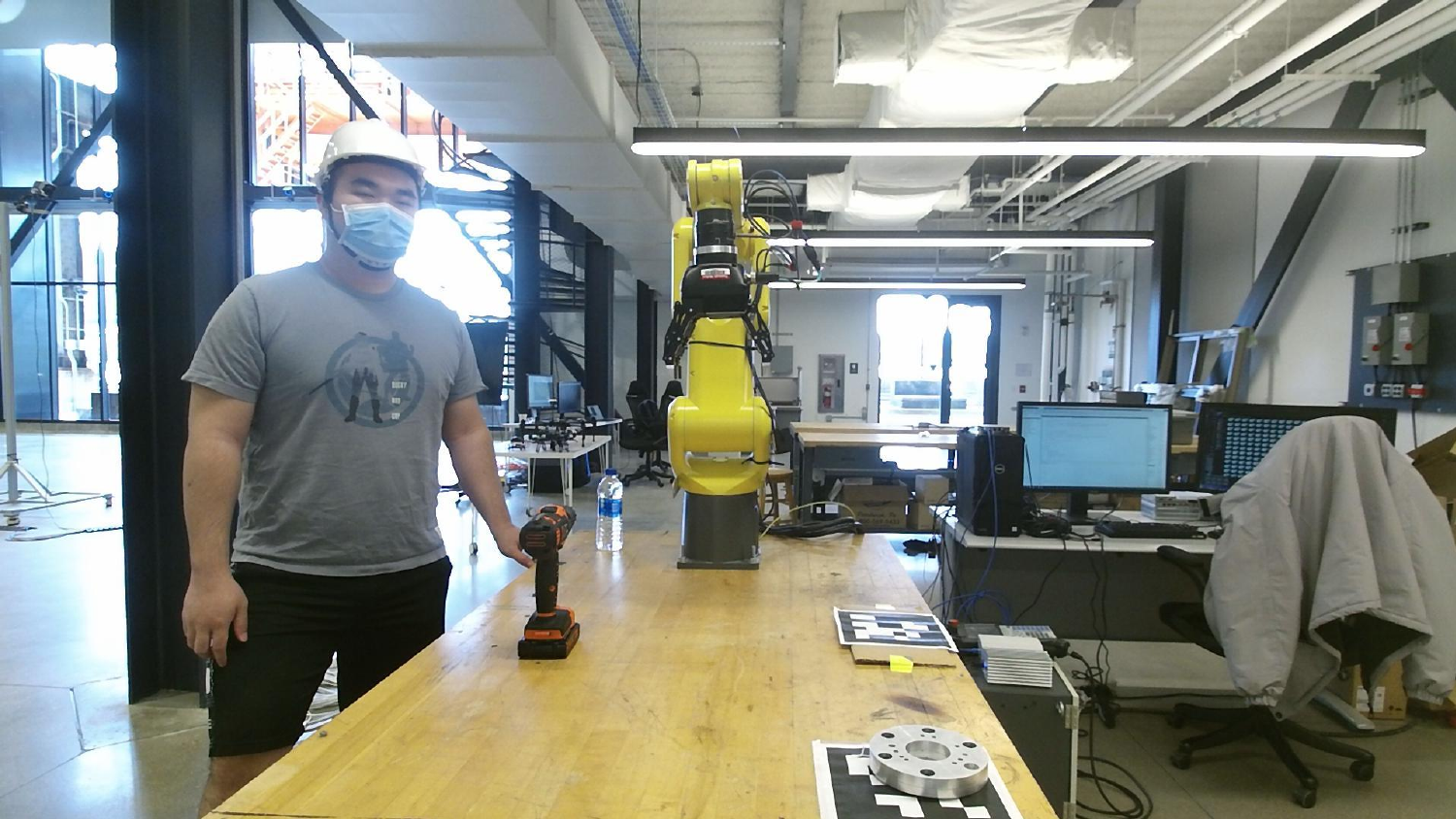}\label{fig:handover_img3}}
\vspace{-10pt}
\caption{Example images in the human-robot handover task.
\label{fig:handover_imgs}}
\vspace{-10pt}
\end{figure}

We also apply IADA to a human-robot handover task (\cref{fig:handover}), where the robot delivers a human-desired tool to the human.
It is important to correctly infer the human intention during the collaboration when the human wants the tool, so that the robot can deliver it to the human in a timely manner.
To study the IADA performance in high-dimensional problems, the input to the classification model is the down-sampled image captured by a Kinect V2 camera, and the output is an intention label indicating whether the human requests the robot to hand over the tool or not.
\Cref{fig:handover_imgs} displays several example raw images (higher resolution) in the dataset.
The training data is $6.25\%$ of the entire dataset, which is limited.
In addition, only $10\%$ of the training data has the human on the left side of the robot, which indicates the training set is imbalanced.
The classifier is constructed using a convolution neural network (CNN) structure, which includes two convolution layers followed by two FCNN layers.
The convolution layers have kernel size 5, stride 2, and padding 1.
The FCNN layers have hidden size 1152 followed by 32.
The learning rate is set to $0.01$ for all methods and the $\epsilon$ in \cref{eq:verification_obj} is set to $0.1$ for IADA.
For IADA, the inner optimization rate is set to be $r = 100$, and we have $C=300$ and $d=0.01$.
Note that for COAP, we set $\epsilon = 10^{-6}$ since the performance deteriorated as $\epsilon$ increased and failed with larger $\epsilon$.
This also indicates the importance of expert guidance in IADA.
The neural network verification is implemented using the $\alpha$, $\beta$-CROWN \cite{zhang2018efficient,xu2021fast,wang2021beta} since it is more capable to handle larger network structures. 

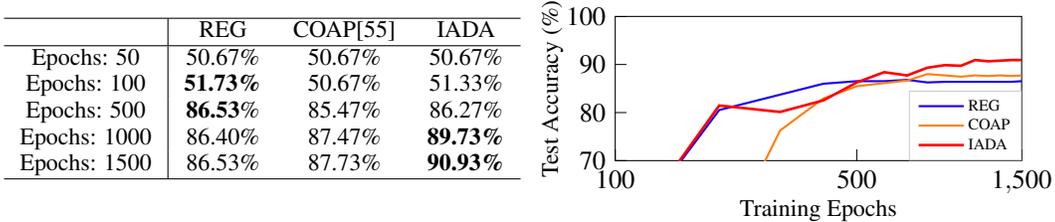
\begin{figure}
\begin{minipage}{0.5\linewidth}
\centering\small
\begin{tabular}{c | c  c  c} 
\hline
 & REG & COAP\cite{convdual} & IADA \\  
\hline
Epochs: 50 &  50.67\% & 50.67\% & 50.67\% \\
Epochs: 100 &  \textbf{51.73\%} & 50.67\% & 51.33\% \\
Epochs: 500 &  \textbf{86.53}\% & 85.47\% & 86.27\% \\
Epochs: 1000 &  86.40\% & 87.47\% & \textbf{89.73\%} \\
Epochs: 1500 &  86.53\% & 87.73\% & \textbf{90.93\%} \\
\hline
\end{tabular}
\vspace{+20pt}
\end{minipage}
\begin{minipage}{0.5\linewidth}
    \centering
    \input{plots/intention2_accuracy}
\end{minipage}
\vspace{-20pt}
\caption{Comparison of the intention prediction accuracy in human-robot handover. Left: numerical values of the model accuracy at selected epochs. Right: model accuracy throughout the training. \label{fig:intention2_accuracy}}
\vspace{-20pt}
\end{figure}

\Cref{fig:intention2_accuracy} demonstrates the performance of the classification models learned using different methods.
The table indicates the model accuracy at epochs $50$, $100$, $500$, $1000$ and $1500$.
We observe that all methods have similar performance at the start at epoch $50$.
REG has a faster start with the best performances at epochs $100$ and $500$. And then quickly reaches a ceiling and does not improve further.
On the other hand, COAP suffers a slow start but passes REG later with better accuracy.
Similarly, IADA also has a slow start but eventually achieves the best performance among the comparing methods.

The plot on the right shows the detailed training performance.
All training methods steadily improve the performance in the first $500$ epochs.
REG quickly converges and stops improving.
Since COAP enlarges the robust boundary around the training data, it steadily improves the accuracy and reaches a better performance compared to REG.
However, it still converges to a ceiling since it only enlarges the decision boundary around the training data, while IADA iteratively expands in the entire data space $X$.
Thus, we observe that IADA keeps improving, though slowly, the model performance due to the iterative data expansion and expert guidance.
Moreover, an increasing trend of the red curve is observed when the training ends at $1500$ epochs, which indicates a further improvement is possible using IADA.

Although the training performance is promising, the training time and required human effort is significantly more compared to other methods.
More discussion on the results (\ie computation efficiency) is included in the supplementary.
In addition, more examples of the training and augmented image data are included in the supplementary.



\section{Conclusion and Future Work} \label{conclusion}
This paper proposes an iterative adversarial data augmentation (IADA) framework to train an NN classifier for intention prediction. It uses neural network verification online to find the most vulnerable part of the network, such as samples on the decision boundary of the network (the most friendly adversaries).
By acquiring human expert knowledge, the framework augments the training data using the adversaries verified by humans and iteratively expands the available data. The experiments demonstrate that our training method can improve the robustness and accuracy of the learned model.

There are many future directions that we would like to pursue, including
exploring efficient online verification algorithms to improve the scalability of the IADA framework; designing a better scheduler in order to balance the required human effort and the learning accuracy; extending the IADA framework to general regression NN (\eg motion prediction) where we will augment new data at places with larger gradients.

\bibliography{asme2e}

\end{document}

%% file: plots/intention2_accuracy.tex
\begin{tikzpicture}

\definecolor{color0}{rgb}{0.1,0,0.95}
\definecolor{color1}{rgb}{1,0,0}
\definecolor{color2}{rgb}{1,0.498,0.0549}
\pgfplotsset{compat=1.5}
\begin{groupplot}[group style={group size=1 by 1}]
\nextgroupplot[
width = \linewidth,
height = 3.5cm,
font = \normalsize,
legend cell align={left},
legend style={nodes={scale=0.6, transform shape}, fill opacity=0.8, draw opacity=1, text opacity=1, draw=white!80!black,at={(1,0)},anchor=south east},
tick align=inside,
tick pos=left,
xmin=100, xmax=1500,
xtick={100, 500, 1500},
xtick style={color=black},
xlabel={Training Epochs},
log ticks with fixed point,
xmode=log,
ylabel={Test Accuracy (\%)},
ymin=70, ymax=100,
ytick={70, 80, 90, 100},
ylabel style={yshift=-0.2cm,font=\small},
xlabel style={yshift=0.15cm,font=\small},
ytick style={color=black},
]
\addplot [thick, color0]
table {%
50	50.67
100	51.73
200	80.53
300	83.73
400	86.00
500	86.53
600	86.53
700	86.80
800	86.27
900	86.40
1000	86.40
1100	86.40
1200	86.40
1300	86.40
1400	86.40
1500	86.53
};
\addlegendentry{REG}

\addplot [thick, color2]
table {%
50	50.67
100	50.67
200	50.67
300	76.27
400	82.93
500	85.47
600	86.13
700	86.67
800	88.00
900	87.73
1000	87.47
1100	87.73
1200	87.60
1300	87.73
1400	87.60
1500	87.73
};
\addlegendentry{COAP}

\addplot [line width=1.0pt, color1]
table {
50	50.67
100	51.33
200	81.47
300	80.13
400	82.53
500	86.27
600	88.40
700	87.73
800	89.33
900	89.87
1000	89.73
1100	90.93
1200	90.67
1300	90.80
1400	90.93
1500	90.93
};
\addlegendentry{IADA}

\end{groupplot}
\end{tikzpicture}

%% file: main.bbl
\begin{thebibliography}{63}
\providecommand{\natexlab}[1]{#1}
\providecommand{\url}[1]{\texttt{#1}}
\expandafter\ifx\csname urlstyle\endcsname\relax
  \providecommand{\doi}[1]{doi: #1}\else
  \providecommand{\doi}{doi: \begingroup \urlstyle{rm}\Url}\fi

\bibitem[Abuduweili et~al.(2019)Abuduweili, Li, and Liu]{abu_intention}
Abulikemu Abuduweili, Siyan Li, and Changliu Liu.
\newblock Adaptable human intention and trajectory prediction for human-robot
  collaboration.
\newblock \emph{arXiv preprint arXiv:1909.05089}, 2019.

\bibitem[Agarap(2018)]{relu}
Abien~Fred Agarap.
\newblock Deep learning using rectified linear units (relu).
\newblock \emph{arXiv preprint arXiv:11803.08375}, 2018.

\bibitem[Anand et~al.(1993)Anand, Mehrotra, Mohan, and Ranka]{286891}
R.~Anand, K.G. Mehrotra, C.K. Mohan, and S.~Ranka.
\newblock An improved algorithm for neural network classification of imbalanced
  training sets.
\newblock \emph{IEEE Transactions on Neural Networks}, 4\penalty0 (6):\penalty0
  962--969, 1993.

\bibitem[Angluin(1988)]{angluin1988queries}
Dana Angluin.
\newblock Queries and concept learning.
\newblock \emph{Machine learning}, 2\penalty0 (4):\penalty0 319--342, 1988.

\bibitem[Angluin(2004)]{ANGLUIN2004175}
Dana Angluin.
\newblock Queries revisited.
\newblock \emph{Theoretical Computer Science}, 313\penalty0 (2):\penalty0
  175--194, 2004.
\newblock Algorithmic Learning Theory.

\bibitem[Antoniou et~al.(2018)Antoniou, Storkey, and Edwards]{antoniou2018data}
Antreas Antoniou, Amos Storkey, and Harrison Edwards.
\newblock Data augmentation generative adversarial networks.
\newblock \emph{arXiv preprint arXiv:1711.04340}, 2018.

\bibitem[Bai et~al.(2021)Bai, Luo, Zhao, Wen, and Wang]{bai2021recent}
Tao Bai, Jinqi Luo, Jun Zhao, Bihan Wen, and Qian Wang.
\newblock Recent advances in adversarial training for adversarial robustness.
\newblock \emph{arXiv preprint arXiv:2102.01356}, 2021.

\bibitem[Bowles et~al.(2018)Bowles, Chen, Guerrero, Bentley, Gunn, Hammers,
  Dickie, Hernández, Wardlaw, and Rueckert]{bowles2018gan}
Christopher Bowles, Liang Chen, Ricardo Guerrero, Paul Bentley, Roger Gunn,
  Alexander Hammers, David~Alexander Dickie, Maria~Valdés Hernández, Joanna
  Wardlaw, and Daniel Rueckert.
\newblock Gan augmentation: Augmenting training data using generative
  adversarial networks.
\newblock \emph{arXiv preprint arXiv:1810.10863}, 2018.

\bibitem[{Cao} et~al.(2019){Cao}, {Hidalgo Martinez}, {Simon}, {Wei}, and
  {Sheikh}]{8765346}
Z.~{Cao}, G.~{Hidalgo Martinez}, T.~{Simon}, S.~{Wei}, and Y.~A. {Sheikh}.
\newblock Openpose: Realtime multi-person 2d pose estimation using part
  affinity fields.
\newblock \emph{IEEE Transactions on Pattern Analysis and Machine
  Intelligence}, 2019.

\bibitem[Cheng et~al.(2020)Cheng, Lei, Chen, Dhillon, and Hsieh]{cheng2020cat}
Minhao Cheng, Qi~Lei, Pin-Yu Chen, Inderjit Dhillon, and Cho-Jui Hsieh.
\newblock Cat: Customized adversarial training for improved robustness.
\newblock \emph{arXiv preprint arXiv:2002.06789}, 2020.

\bibitem[{Cheng} et~al.(2019){Cheng}, {Zhao}, {Liu}, and {Tomizuka}]{8814980}
Y.~{Cheng}, W.~{Zhao}, C.~{Liu}, and M.~{Tomizuka}.
\newblock Human motion prediction using semi-adaptable neural networks.
\newblock In \emph{2019 American Control Conference (ACC)}, pages 4884--4890,
  2019.

\bibitem[Cohn et~al.(1994)Cohn, Atlas, and Ladner]{10.1023/A:1022673506211}
David Cohn, Les Atlas, and Richard Ladner.
\newblock Improving generalization with active learning.
\newblock \emph{Mach. Learn.}, 15\penalty0 (2):\penalty0 201–221, May 1994.

\bibitem[Cubuk et~al.(2019)Cubuk, Zoph, Mane, Vasudevan, and
  Le]{cubuk2019autoaugment}
Ekin~D. Cubuk, Barret Zoph, Dandelion Mane, Vijay Vasudevan, and Quoc~V. Le.
\newblock Autoaugment: Learning augmentation policies from data.
\newblock \emph{arXiv preprint arXiv:1805.09501}, 2019.

\bibitem[DeVries and Taylor(2017)]{cutout}
Terrance DeVries and Graham~W. Taylor.
\newblock Improved regularization of convolutional neural networks with cutout.
\newblock \emph{arXiv preprint arXiv:1708.04552}, 2017.

\bibitem[Ducoffe and Precioso(2018)]{ducoffe2018adversarial}
Melanie Ducoffe and Frederic Precioso.
\newblock Adversarial active learning for deep networks: a margin based
  approach.
\newblock \emph{arXiv preprint arXiv:1802.09841}, 2018.

\bibitem[Elsayed et~al.(2018)Elsayed, Shankar, Cheung, Papernot, Kurakin,
  Goodfellow, and Sohl-Dickstein]{NEURIPS2018_8562ae5e}
Gamaleldin Elsayed, Shreya Shankar, Brian Cheung, Nicolas Papernot, Alexey
  Kurakin, Ian Goodfellow, and Jascha Sohl-Dickstein.
\newblock Adversarial examples that fool both computer vision and time-limited
  humans.
\newblock In S.~Bengio, H.~Wallach, H.~Larochelle, K.~Grauman, N.~Cesa-Bianchi,
  and R.~Garnett, editors, \emph{Advances in Neural Information Processing
  Systems}, volume~31. Curran Associates, Inc., 2018.

\bibitem[Gong et~al.(2021)Gong, Ren, Ye, and Liu]{maxup}
Chengyue Gong, Tongzheng Ren, Mao Ye, and Qiang Liu.
\newblock Maxup: Lightweight adversarial training with data augmentation
  improves neural network training.
\newblock In \emph{2021 IEEE/CVF Conference on Computer Vision and Pattern
  Recognition (CVPR)}, pages 2474--2483, 2021.

\bibitem[Goodfellow et~al.(2015)Goodfellow, Shlens, and
  Szegedy]{goodfellow2015explaining}
Ian Goodfellow, Jonathon Shlens, and Christian Szegedy.
\newblock Explaining and harnessing adversarial examples.
\newblock In \emph{International Conference on Learning Representations}, 2015.

\bibitem[He and Garcia(2009)]{5128907}
Haibo He and Edwardo~A. Garcia.
\newblock Learning from imbalanced data.
\newblock \emph{IEEE Transactions on Knowledge and Data Engineering},
  21\penalty0 (9):\penalty0 1263--1284, 2009.

\bibitem[Hendrycks et~al.(2021)Hendrycks, Basart, Mu, Kadavath, Wang, Dorundo,
  Desai, Zhu, Parajuli, Guo, Song, Steinhardt, and Gilmer]{deepaug}
D.~Hendrycks, S.~Basart, N.~Mu, S.~Kadavath, F.~Wang, E.~Dorundo, R.~Desai,
  T.~Zhu, S.~Parajuli, M.~Guo, D.~Song, J.~Steinhardt, and J.~Gilmer.
\newblock The many faces of robustness: A critical analysis of
  out-of-distribution generalization.
\newblock In \emph{2021 IEEE/CVF International Conference on Computer Vision
  (ICCV)}, pages 8320--8329, Los Alamitos, CA, USA, oct 2021. IEEE Computer
  Society.

\bibitem[Hendrycks et~al.(2020)Hendrycks, Mu, Cubuk, Zoph, Gilmer, and
  Lakshminarayanan]{augmix}
Dan Hendrycks, Norman Mu, Ekin~D. Cubuk, Barret Zoph, Justin Gilmer, and Balaji
  Lakshminarayanan.
\newblock Augmix: A simple data processing method to improve robustness and
  uncertainty.
\newblock \emph{Proceedings of the International Conference on Learning
  Representations (ICLR)}, 2020.

\bibitem[Japkowicz and Stephen(2002)]{Japkowicz02theclass}
Nathalie Japkowicz and Shaju Stephen.
\newblock The class imbalance problem: A systematic study.
\newblock \emph{Intelligent Data Analysis}, pages 429--449, 2002.

\bibitem[Kingma and Ba(2017)]{kingma2017adam}
Diederik~P. Kingma and Jimmy Ba.
\newblock Adam: A method for stochastic optimization.
\newblock \emph{arXiv preprint arXiv:1412.6980}, 2017.

\bibitem[Krawczyk(2016)]{krawczyk2016learning}
Bartosz Krawczyk.
\newblock Learning from imbalanced data: open challenges and future directions.
\newblock \emph{Progress in Artificial Intelligence}, 5\penalty0 (4):\penalty0
  221--232, 2016.

\bibitem[Lasota et~al.(2017)Lasota, Song, and Shah]{book}
P.A. Lasota, T.~Song, and J.A. Shah.
\newblock \emph{A Survey of Methods for Safe Human-Robot Interaction}.
\newblock Foundations and Trends(r) in Robotics Series. Now Publishers, 2017.

\bibitem[Lemley et~al.(2017)Lemley, Bazrafkan, and Corcoran]{Lemley_2017}
Joseph Lemley, Shabab Bazrafkan, and Peter Corcoran.
\newblock Smart augmentation learning an optimal data augmentation strategy.
\newblock \emph{IEEE Access}, 5:\penalty0 5858–5869, 2017.

\bibitem[Lewis and Gale(1994)]{10.1007/978-1-4471-2099-5_1}
David~D. Lewis and William~A. Gale.
\newblock A sequential algorithm for training text classifiers.
\newblock In Bruce~W. Croft and C.~J. van Rijsbergen, editors, \emph{SIGIR
  '94}, pages 3--12, London, 1994. Springer London.

\bibitem[Li et~al.(2018)Li, Zhang, and Diao]{8525781}
Shengchao Li, Lin Zhang, and Xiumin Diao.
\newblock Improving human intention prediction using data augmentation.
\newblock In \emph{2018 27th IEEE International Symposium on Robot and Human
  Interactive Communication (RO-MAN)}, pages 559--564, 2018.

\bibitem[Liu et~al.(2019)Liu, Arnon, Lazarus, and Kochenderfer]{Liu2019iclr}
Changliu Liu, Tomer Arnon, Christopher Lazarus, and Mykel~J. Kochenderfer.
\newblock {N}eural{V}erification.jl: {A}lgorithms for verifying deep neural
  networks.
\newblock In \emph{Workshop on Debugging Machine Learning,}, 2019.

\bibitem[Liu et~al.(2021)Liu, Arnon, Lazarus, Strong, Barrett, and
  Kochenderfer]{liu2020algorithms}
Changliu Liu, Tomer Arnon, Christopher Lazarus, Christopher Strong, Clark
  Barrett, and Mykel~J. Kochenderfer.
\newblock Algorithms for verifying deep neural networks.
\newblock \emph{Foundations and Trends® in Optimization}, 4\penalty0
  (3-4):\penalty0 244--404, 2021.

\bibitem[Liu and Liu(2021)]{9281312}
Ruixuan Liu and Changliu Liu.
\newblock Human motion prediction using adaptable recurrent neural networks and
  inverse kinematics.
\newblock \emph{IEEE Control Systems Letters}, 5\penalty0 (5):\penalty0
  1651--1656, 2021.

\bibitem[Liu et~al.(2022{\natexlab{a}})Liu, Chen, and Liu]{jssa}
Ruixuan Liu, Rui Chen, and Changliu Liu.
\newblock Safe interactive industrial robots using jerk-based safe set
  algorithm.
\newblock \emph{arXiv preprint arXiv:2204.03038}, 2022{\natexlab{a}}.

\bibitem[Liu et~al.(2022{\natexlab{b}})Liu, Chen, Sun, Zhao, and Liu]{jpc}
Ruixuan Liu, Rui Chen, Yifan Sun, Yu~Zhao, and Changliu Liu.
\newblock Jerk-bounded position controller with real-time task modification for
  interactive industrial robots.
\newblock 1 2022{\natexlab{b}}.

\bibitem[Lomuscio and Maganti(2017)]{lomuscio2017approach}
Alessio Lomuscio and Lalit Maganti.
\newblock An approach to reachability analysis for feed-forward relu neural
  networks.
\newblock \emph{arXiv preprint arXiv:1706.07351}, 2017.

\bibitem[Lopes et~al.(2019)Lopes, Yin, Poole, Gilmer, and Cubuk]{da_gaussian}
Raphael~Gontijo Lopes, Dong Yin, Ben Poole, Justin Gilmer, and Ekin~D. Cubuk.
\newblock Improving robustness without sacrificing accuracy with patch gaussian
  augmentation.
\newblock \emph{arXiv preprint arXiv:1906.02611}, 2019.

\bibitem[Matheson et~al.(2019)Matheson, Minto, Zampieri, Faccio, and
  Rosati]{Matheson_2019}
Eloise Matheson, Riccardo Minto, Emanuele G.~G. Zampieri, Maurizio Faccio, and
  Giulio Rosati.
\newblock Human–robot collaboration in manufacturing applications: A review.
\newblock \emph{Robotics}, 8\penalty0 (4):\penalty0 100, Dec 2019.

\bibitem[Mayer and Timofte(2020)]{Mayer_2020_WACV}
Christoph Mayer and Radu Timofte.
\newblock Adversarial sampling for active learning.
\newblock In \emph{Proceedings of the IEEE/CVF Winter Conference on
  Applications of Computer Vision (WACV)}, March 2020.

\bibitem[Moosavi-Dezfooli et~al.(2016)Moosavi-Dezfooli, Fawzi, and
  Frossard]{moosavidezfooli2016deepfool}
S.~Moosavi-Dezfooli, A.~Fawzi, and P.~Frossard.
\newblock Deepfool: A simple and accurate method to fool deep neural networks.
\newblock In \emph{2016 IEEE Conference on Computer Vision and Pattern
  Recognition (CVPR)}, pages 2574--2582, Los Alamitos, CA, USA, jun 2016. IEEE
  Computer Society.

\bibitem[Paszke et~al.(2019)Paszke, Gross, Massa, Lerer, Bradbury, Chanan,
  Killeen, Lin, Gimelshein, Antiga, Desmaison, Kopf, Yang, DeVito, Raison,
  Tejani, Chilamkurthy, Steiner, Fang, Bai, and Chintala]{NEURIPS2019_9015}
Adam Paszke, Sam Gross, Francisco Massa, Adam Lerer, James Bradbury, Gregory
  Chanan, Trevor Killeen, Zeming Lin, Natalia Gimelshein, Luca Antiga, Alban
  Desmaison, Andreas Kopf, Edward Yang, Zachary DeVito, Martin Raison, Alykhan
  Tejani, Sasank Chilamkurthy, Benoit Steiner, Lu~Fang, Junjie Bai, and Soumith
  Chintala.
\newblock Pytorch: An imperative style, high-performance deep learning library.
\newblock In \emph{Advances in Neural Information Processing Systems 32}, pages
  8024--8035. 2019.

\bibitem[Perez and Wang(2017)]{perez2017effectiveness}
Luis Perez and Jason Wang.
\newblock The effectiveness of data augmentation in image classification using
  deep learning.
\newblock \emph{arXiv preprint arXiv:1712.04621}, 2017.

\bibitem[Phillips et~al.(2017)Phillips, Wheeler, and Kochenderfer]{7995948}
Derek~J. Phillips, Tim~A. Wheeler, and Mykel~J. Kochenderfer.
\newblock Generalizable intention prediction of human drivers at intersections.
\newblock In \emph{2017 IEEE Intelligent Vehicles Symposium (IV)}, pages
  1665--1670, 2017.

\bibitem[Ross et~al.(2011)Ross, Gordon, and Bagnell]{dagger}
Stephane Ross, Geoffrey Gordon, and Drew Bagnell.
\newblock A reduction of imitation learning and structured prediction to
  no-regret online learning.
\newblock In \emph{Proceedings of the Fourteenth International Conference on
  Artificial Intelligence and Statistics}, volume~15 of \emph{Proceedings of
  Machine Learning Research}, pages 627--635. PMLR, 11--13 Apr 2011.

\bibitem[Rudenko et~al.(2020)Rudenko, Palmieri, Herman, Kitani, Gavrila, and
  Arras]{rudenko}
Andrey Rudenko, Luigi Palmieri, Michael Herman, Kris~M Kitani, Dariu~M Gavrila,
  and Kai~O Arras.
\newblock Human motion trajectory prediction: a survey.
\newblock \emph{The International Journal of Robotics Research}, 39\penalty0
  (8):\penalty0 895--935, 2020.

\bibitem[Settles(2010)]{settles2009active}
Burr Settles.
\newblock Active learning literature survey.
\newblock \emph{University of Wisconsin, Madison}, 52, 07 2010.

\bibitem[Settles et~al.(2008)Settles, Craven, and Ray]{NIPS2007_a1519de5}
Burr Settles, Mark Craven, and Soumya Ray.
\newblock Multiple-instance active learning.
\newblock In J.~Platt, D.~Koller, Y.~Singer, and S.~Roweis, editors,
  \emph{Advances in Neural Information Processing Systems}, volume~20. Curran
  Associates, Inc., 2008.

\bibitem[Seung et~al.(1992)Seung, Opper, and
  Sompolinsky]{10.1145/130385.130417}
H.~S. Seung, M.~Opper, and H.~Sompolinsky.
\newblock Query by committee.
\newblock In \emph{Proceedings of the Fifth Annual Workshop on Computational
  Learning Theory}, COLT '92, page 287–294, New York, NY, USA, 1992.
  Association for Computing Machinery.

\bibitem[Shorten and Khoshgoftaar(2019)]{Shorten2019ASO}
Connor Shorten and T.~Khoshgoftaar.
\newblock A survey on image data augmentation for deep learning.
\newblock \emph{Journal of Big Data}, 6:\penalty0 1--48, 2019.

\bibitem[Szegedy et~al.(2014)Szegedy, Zaremba, Sutskever, Bruna, Erhan,
  Goodfellow, and Fergus]{szegedy2014intriguing}
Christian Szegedy, Wojciech Zaremba, Ilya Sutskever, Joan Bruna, Dumitru Erhan,
  Ian Goodfellow, and Rob Fergus.
\newblock Intriguing properties of neural networks.
\newblock In \emph{International Conference on Learning Representations}, 2014.

\bibitem[Tjeng et~al.(2019)Tjeng, Xiao, and Tedrake]{tjeng2019evaluating}
Vincent Tjeng, Kai Xiao, and Russ Tedrake.
\newblock Evaluating robustness of neural networks with mixed integer
  programming.
\newblock \emph{arXiv preprint arXiv:1711.07356}, 2019.

\bibitem[Tsipras et~al.(2019)Tsipras, Santurkar, Engstrom, Turner, and
  Madry]{tsipras2019robustness}
Dimitris Tsipras, Shibani Santurkar, Logan Engstrom, Alexander Turner, and
  Aleksander Madry.
\newblock Robustness may be at odds with accuracy.
\newblock \emph{arXiv preprint arXiv:1805.12152}, 2019.

\bibitem[Villani et~al.(2018)Villani, Pini, Leali, and Secchi]{VILLANI2018248}
Valeria Villani, Fabio Pini, Francesco Leali, and Cristian Secchi.
\newblock Survey on human–robot collaboration in industrial settings: Safety,
  intuitive interfaces and applications.
\newblock \emph{Mechatronics}, 55:\penalty0 248 -- 266, 2018.

\bibitem[Wang et~al.(2017)Wang, Zhang, Li, Zhang, and Lin]{Wang_2017}
Keze Wang, Dongyu Zhang, Ya~Li, Ruimao Zhang, and Liang Lin.
\newblock Cost-effective active learning for deep image classification.
\newblock \emph{IEEE Transactions on Circuits and Systems for Video
  Technology}, 27\penalty0 (12):\penalty0 2591–2600, Dec 2017.

\bibitem[Wang et~al.(2021)Wang, Zhang, Xu, Lin, Jana, Hsieh, and
  Kolter]{wang2021beta}
Shiqi Wang, Huan Zhang, Kaidi Xu, Xue Lin, Suman Jana, Cho-Jui Hsieh, and
  J~Zico Kolter.
\newblock {Beta-CROWN}: Efficient bound propagation with per-neuron split
  constraints for complete and incomplete neural network verification.
\newblock \emph{Advances in Neural Information Processing Systems}, 34, 2021.

\bibitem[Weng et~al.(2018)Weng, Zhang, Chen, Song, Hsieh, Boning, Dhillon, and
  Daniel]{weng2018fast}
Tsui-Wei Weng, Huan Zhang, Hongge Chen, Zhao Song, Cho-Jui Hsieh, Duane Boning,
  Inderjit~S. Dhillon, and Luca Daniel.
\newblock Towards fast computation of certified robustness for relu networks.
\newblock \emph{arXiv preprint arXiv:1804.09699}, 2018.

\bibitem[Wong and Kolter(2018)]{convdual}
Eric Wong and Zico Kolter.
\newblock Provable defenses against adversarial examples via the convex outer
  adversarial polytope.
\newblock In \emph{Proceedings of the 35th International Conference on Machine
  Learning}, volume~80 of \emph{Proceedings of Machine Learning Research},
  pages 5286--5295. PMLR, 10--15 Jul 2018.

\bibitem[Xie et~al.(2020)Xie, Tan, Gong, Wang, Yuille, and Le]{advaug}
C.~Xie, M.~Tan, B.~Gong, J.~Wang, A.~L. Yuille, and Q.~V. Le.
\newblock Adversarial examples improve image recognition.
\newblock In \emph{2020 IEEE/CVF Conference on Computer Vision and Pattern
  Recognition (CVPR)}, pages 816--825, Los Alamitos, CA, USA, jun 2020. IEEE
  Computer Society.

\bibitem[Xu et~al.(2021)Xu, Zhang, Wang, Wang, Jana, Lin, and
  Hsieh]{xu2021fast}
Kaidi Xu, Huan Zhang, Shiqi Wang, Yihan Wang, Suman Jana, Xue Lin, and Cho-Jui
  Hsieh.
\newblock {Fast and Complete}: Enabling complete neural network verification
  with rapid and massively parallel incomplete verifiers.
\newblock In \emph{International Conference on Learning Representations}, 2021.

\bibitem[Zhang et~al.(2018{\natexlab{a}})Zhang, Cisse, Dauphin, and
  Lopez-Paz]{mixup}
Hongyi Zhang, Moustapha Cisse, Yann~N. Dauphin, and David Lopez-Paz.
\newblock mixup: Beyond empirical risk minimization.
\newblock In \emph{International Conference on Learning Representations},
  2018{\natexlab{a}}.

\bibitem[Zhang et~al.(2018{\natexlab{b}})Zhang, Weng, Chen, Hsieh, and
  Daniel]{zhang2018efficient}
Huan Zhang, Tsui-Wei Weng, Pin-Yu Chen, Cho-Jui Hsieh, and Luca Daniel.
\newblock Efficient neural network robustness certification with general
  activation functions.
\newblock \emph{Advances in Neural Information Processing Systems},
  31:\penalty0 4939--4948, 2018{\natexlab{b}}.

\bibitem[Zhang et~al.(2020{\natexlab{a}})Zhang, Xu, Han, Niu, Cui, Sugiyama,
  and Kankanhalli]{zhang2020attacks}
Jingfeng Zhang, Xilie Xu, Bo~Han, Gang Niu, Lizhen Cui, Masashi Sugiyama, and
  Mohan Kankanhalli.
\newblock Attacks which do not kill training make adversarial learning
  stronger.
\newblock In Hal~Daumé III and Aarti Singh, editors, \emph{Proceedings of the
  37th International Conference on Machine Learning}, volume 119 of
  \emph{Proceedings of Machine Learning Research}, pages 11278--11287. PMLR,
  13--18 Jul 2020{\natexlab{a}}.

\bibitem[Zhang et~al.(2020{\natexlab{b}})Zhang, Wang, Zhang, and
  Zhong]{Zhang2020AdversarialA}
Xinyu Zhang, Qiang Wang, Jian Zhang, and Zhaobai Zhong.
\newblock Adversarial autoaugment.
\newblock \emph{arXiv preprint arXiv:1912.11188}, 2020{\natexlab{b}}.

\bibitem[Zhong et~al.(2020)Zhong, Zheng, Kang, Li, and Yang]{zhong2017random}
Zhun Zhong, Liang Zheng, Guoliang Kang, Shaozi Li, and Yi~Yang.
\newblock Random erasing data augmentation.
\newblock \emph{Proceedings of the AAAI Conference on Artificial Intelligence},
  34\penalty0 (07):\penalty0 13001--13008, Apr. 2020.

\bibitem[Zhu and Bento(2017)]{zhu2017generative}
Jia-Jie Zhu and José Bento.
\newblock Generative adversarial active learning.
\newblock \emph{arXiv preprint arXiv:1702.07956}, 2017.

\end{thebibliography}
